\documentclass[preprint,12pt]{elsarticle}

\usepackage{amssymb}
\usepackage{amsmath}

\usepackage{booktabs}
\usepackage{wrapfig}   
\usepackage{float}     
\usepackage{graphicx}  
\usepackage{caption}   
\journal{Nuclear Physics B}

\begin{document}

\begin{frontmatter}



\title{Revisiting Direct Encoding: Learnable Temporal Dynamics for Static Image Spiking Neural Networks}


\author[1,2]{Huaxu He\corref{cor1}}

\affiliation[1]{
  organization={School of Computer and Information Engineering, Henan University},
  city={Kaifeng},
  postcode={475004}, 
  state={Henan},
  country={China}
}
\affiliation[2]{
  organization={ Henan Key Laboratory of Big Data Analysis and Processing, Henan University},
   city={Kaifeng},
  postcode={475004}, 
  state={Henan},
  country={China}
}

\cortext[cor1]{Corresponding author: Huaxu He (email: 104753230975@henu.edu.cn) }

\begin{abstract}
Handling static images that lack inherent temporal dynamics remains a fundamental challenge for spiking neural networks (SNNs). In directly trained SNNs, static inputs are typically repeated across time steps, causing the temporal dimension to collapse into a rate like representation and preventing meaningful temporal modeling. This work revisits the reported performance gap between direct and rate based encodings and shows that it primarily stems from convolutional learnability and surrogate gradient formulations rather than the encoding schemes themselves. To illustrate this mechanism level clarification, we introduce a minimal learnable temporal encoding that adds adaptive phase shifts to induce meaningful temporal variation from static inputs.
\end{abstract}




\end{frontmatter}


\section{Introduction}
\label{sec1}

Spiking neural networks (SNNs), with event driven sparsity and spatiotemporal processing, offer brain inspired and energy efficient computation \cite{2,3,4,5,6}. However, in static image tasks, existing SNNs still rely on approximate schemes such as rate coding, leaving their temporal modeling capability underutilized. Achieving meaningful temporal dynamics under low time-step constraints remains challenging \cite{7}.

Within current SNN research, training is broadly divided into ANN-to-SNN conversion and direct training \cite{8,10,11,24}.
Conversion methods approximate ANN activations with firing rates and typically require many time steps \cite{13}.
while direct training allows end-to-end learning with just a few steps \cite{9}.
Among direct methods, direct encoding is the most representative, enabling efficient inference at minimal time steps.

We revisit the widely adopted direct encoding mechanism in directly trained SNNs.
Although direct encoding enables efficient inference at minimal time steps—demonstrating a clear advantage in computational efficiency—it effectively repeats identical inputs across time steps, causing the temporal dimension to lose its genuine dynamic evolution \cite{1}.

Specifically, in directly trained SNNs, static images are typically repeated along the temporal dimension and fed into leaky integrate-and-fire (LIF) neurons to construct a formal time sequence \cite{14}. However, since the inputs at all time steps are identical, the temporal dimension functions merely as a statistical averaging channel rather than a true temporal modeling mechanism. For LIF neurons under constant input current, the spiking behavior becomes simple and stable, implying that the overall network effectively performs rate-based coding \cite{15}, A minimal illustrative example of this equivalence is provided in the~\ref{app2}.


\begin{figure}[hb]  
\centering
\includegraphics[width=0.98\linewidth]{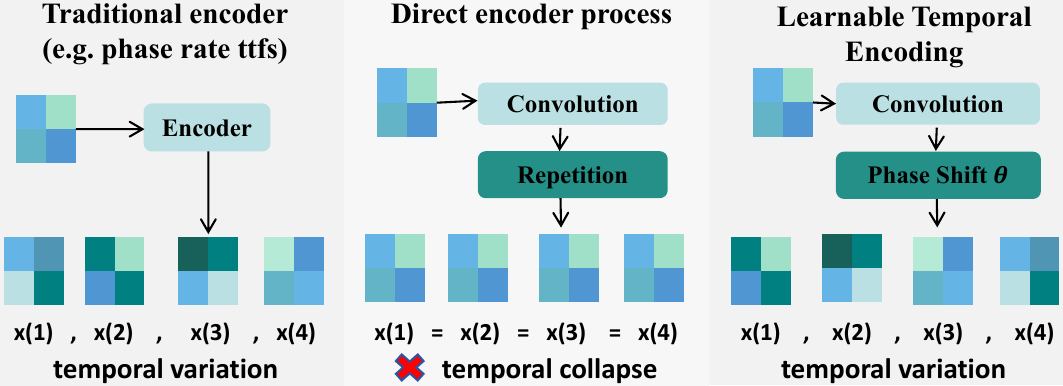}
\caption{ Comparison of traditional encoders, direct encoding, 
and the proposed learnable temporal encoding. Traditional 
encoders produce time-varying inputs via fixed rules, while 
direct encoding replicates static inputs, causing temporal 
collapse. The proposed method restores temporal variation 
through learnable phase shifts. Here, \(  x \) represents the input feature maps at each time step.}\label{fig:1}
\end{figure}

To further analyze the performance discrepancy between direct and rate coding reported in previous studies, we show that the difference mainly arises from the encoding target: rate coding is applied directly to raw inputs, whereas direct encoding operates on static feature maps extracted by convolutional layers. Empirically, the two exhibit highly similar performance in static tasks. The previously reported performance gap is thus largely due to implementation details such as feature selection and surrogate gradient configurations rather than intrinsic differences in their coding principles.

Neurophysiological studies indicate that biological neurons encode information not only through firing rates but also through temporal features like spike timing. Motivated by this, we introduce a learnable temporal encoding that adds trainable phase parameters at the input stage. These parameters induce phase shifts across time steps, forming distinguishable temporal patterns while preserving the structural simplicity of direct encoding. This mechanism equips the network with temporal correlations for learning discriminative representations, as illustrated in Figure \ref{fig:1}.
The main contributions of this work are summarized as follows:

\noindent 

\begin{itemize}
\item We identify the functional equivalence between direct and rate encoding in static image SNNs and show that their reported performance gap mainly results from convolutional learnability and surrogate gradient.
\item We introduce a minimal learnable temporal encoding for static inputs, using trainable phase parameters to induce temporal variation with minimal structural overhead.
\item We validate this design on object recognition and detection tasks using the CIFAR-10, CIFAR-100, and VOC datasets, showing improved performance under low time-step constraints.
\end{itemize}


\section{Related Work}

\subsection{SNN Encoding Schemes}
\label{subsec1}

The encoding problem in SNNs can be categorized into two types:encoding for static images in directly trained SNNs \cite{16}, and feature encoding for intermediate layers in ANN-to-SNN conversion.This work focuses on the former.

In directly trained SNNs, the predominant strategy is direct encoding, which replicates static feature maps along the temporal dimension and feeds them into LIF neurons, producing spike sequences that formally exhibit temporal structure. This approach is simple, stable during training, and performs well under very few time steps.

Other commonly used schemes include rate encoding, time-to-first-spike (TTFS), and phase encoding. Rate encoding conveys signal magnitude through firing rates, whereas TTFS and phase encoding rely on spike timing, with TTFS enforcing a single spike per neuron \cite{17,18,19}.

We show that direct encoding is functionally equivalent to rate encoding, with the key difference being the encoding target:
rate encoding operates directly on raw images,
whereas direct encoding applies to static feature maps produced by convolutional layers.
Based on this observation, we further introduce phase encoding and TTFS encoding
to enhance the temporal representation capability of SNNs.

\subsection{Spiking Neurons}

The LIF neuron model with soft reset,
widely used in current SNNs \cite{20,21},
It can be described as follows:
\begin{equation}
H[t] = \mathcal{L} V[t-1] + (1 - \mathcal{L}) I[t], \quad 0 < \mathcal{L} < 1
\end{equation}
\begin{equation}
S[t] = \Theta \left( H[t] - V_{\text{th}} \right)
\end{equation}
\begin{equation}
V[t] = H[t] - S[t] \times V_{\text{th}}
\end{equation}

Here, \( \mathcal{L} \) denotes the membrane potential decay constant,  
\( I[t] \) is the input current at time step \( t \), and the firing threshold is \( V_{\text{th}} = 1 \).  
A spike \( S[t] \) is emitted when the membrane potential \( H[t] \) exceeds \( V_{\text{th}} \).  
\( \Theta(x) \) represents the Heaviside step function, which equals 1 when \( x \ge 0 \) and 0 otherwise.  
After firing, the membrane potential \( V[t] \) is reset by subtracting \( S[t] \times V_{\text{th}} \);  
if no spike occurs, \( V[t] = H[t] \).

\vspace{6pt}

To better process phase dependent information, we modify the standard LIF neuron as follows:
\begin{equation}
H[t] = \mathcal{L}_t^{\text{learn}} V[t-1] + \beta_t^{\text{learn}} I[t]
\end{equation}

where \( \mathcal{L}_t^{\text{learn}} \) represents a learnable membrane potential decay constant  
at time step \( t \), and \( \beta_t^{\text{learn}} \) denotes a learnable weight  
that controls the contribution of the input current at each time step.

Since the input current can be expressed as \( I[t] = W^{(n-1)} S^{(n-1)}[t] \),  
the modified neuron enables the network to adaptively rescale the spike inputs  
\( S^{(n-1)}[t] \) through the learnable parameters \( \beta_t^{\text{learn}} \).  
Here, \( n \) denotes the layer index and \( W \) represents the convolutional or fully connected weights.  
The learnable parameters are shared among all neurons within each channel.

\section{Method}
\label{sec1}
This section first introduces how to optimize rate coding through learnable convolutional layers and surrogate gradients, enabling comparable performance to direct encoding under very few time steps. Subsequently, a learnable temporal encoding mechanism is proposed to enhance the network’s ability to model temporal dependencies.

\subsection{Learnability of Convolutional Layers and Surrogate Gradient}
\label{subsec1}

In conventional SNN encoding schemes such as rate coding, TTFS, and phase coding a relatively large number of time steps are typically required to faithfully represent all input information (e.g., every pixel in an image). This not only escalates training costs and energy consumption but also inevitably introduces redundant spike activities lacking discriminative power at the input stage, thereby undermining the efficiency of temporal feature representation.

In contrast, direct encoding achieves competitive performance with extremely few time steps by exploiting the \textbf{learnability of preceding convolutional layers}, rather than relying on intrinsic temporal dynamics. These convolutional layers extract discriminative spatial features prior to the spiking transformation.

Motivated by this observation, we prepend convolutional layers to traditional encoding modules, enabling the network to adaptively form task effective spike representations and substantially narrow the performance gap with direct encoding.

However, directly applying binary operations (e.g., rounding or Heaviside step functions) to the convolutional outputs leads to gradient blockage, thereby preventing effective parameter updates in the preceding convolutional layers during backpropagation. To address this issue, a surrogate gradient is employed to smoothly approximate the non differentiable spiking activation, enabling end-to-end optimization \cite{22}. This design significantly enhances feature learning under low time step constraints and provides a solid foundation for subsequent temporal modeling.

\subsection{Learnable Temporal Encoding}
\label{subsec1}

\begin{wrapfigure}{r}{0.58\textwidth}
    \vspace{-0.5\baselineskip}
    \centering
    \includegraphics[width=0.48\textwidth]{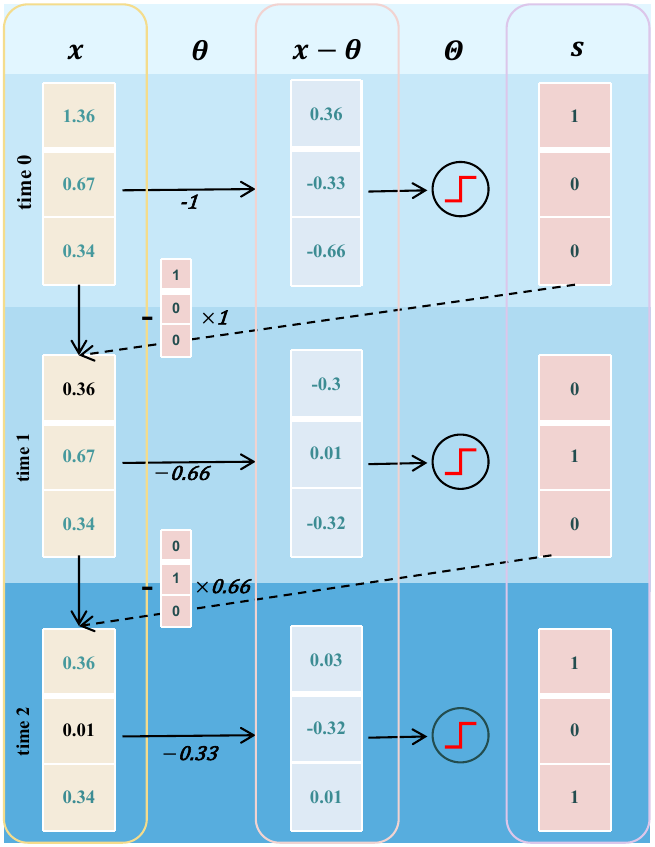}
    \caption{
        Example of the learnable temporal encoding (phase encoding).
        Values $\theta_0 = 1$, $\theta_1 = 0.66$, and $\theta_2 = 0.33$
        are illustrative; actual $\theta_t$ are learnable.
    }
    \label{fig:phase}
    \vspace{-1.0\baselineskip}
\end{wrapfigure}

Building upon the analysis of the temporal degeneration in direct encoding, we propose a learnable temporal encoding mechanism that introduces trainable temporal dependencies under static input conditions. This mechanism transforms the convolutional feature maps into spike sequences with controllable temporal variations. (See Figure 
\ref{fig:phase} for an illustration of the mechanism).

Formally, given an input feature map $X$, it is replicated along the temporal axis to form $\{X_t\}_{t=1}^T$. At each time step, the spike activation is computed as:
\begin{equation}
s_t = \Theta(x_t - \theta_t),
\end{equation}

where $\Theta(\cdot)$ is the Heaviside step function.which is approximated by a surrogate gradient during backpropagation.To emulate temporal progression, the firing threshold is adaptively decayed across time steps according to the following.
\begin{equation}
\theta_{t+1} = \theta_t \times \sigma(a_t),
\end{equation}

The adaptively decaying threshold $\theta_{t}$ (controlled by a learnable, channel-wise shared 
$a_t$)
introduces a time varying activation boundary, enabling the network to learn temporal patterns in spike generation under different encoding rules. After spike generation, different update rules lead to three classical temporal encoding forms:
\paragraph{1)Phase Encoding}
\begin{equation}
x_{t+1} = x_t - s_t \times \theta_t
\end{equation}
Where each spike subtracts the threshold term from the input, producing progressively phase shifted feature sequences.

\paragraph{2)TTFS}
\begin{equation}
x_{t+1} = x_t - s_t \times x_t
\end{equation}

Which enforces that each neuron fires at most once by setting its input to zero after the first spike.
\subsubsection*{3)Rate Encoding.}
When no temporal update is applied, the input features are simply repeated across time, corresponding to conventional rate coding.

This unified framework jointly describes these three spike generation paradigms and enables end-to-end optimization via learnable thresholds and surrogate gradients.
It preserves structural simplicity while providing explicit temporal structures, allowing static inputs to be represented as temporally dependent spike patterns.

\section{Experiments}
\label{sec1}


\begin{table}[b]
\centering
\caption{Ablation study on CIFAR-10 using TTFS.
LT = Learnable Threshold $\theta_{t}$ ,LC = Learnability of Convolution Layer,SG=Surrogate Gradient.}
\setlength{\tabcolsep}{6pt}
\renewcommand{\arraystretch}{1.1}
\begin{tabular}{lcccc}
\hline
\textbf{Method}  & \textbf{LT} & \textbf{LC} & \textbf{SG} & \textbf{Acc@1} \\
\hline
TTFS           &             &              &              & 83.22 \\
TTFS+LT         & $\checkmark$ &              &              & 83.44 \\
TTFS+LT+LC       & $\checkmark$ & $\checkmark$ &              & 94.32 \\
TTFS+LT+LC+SG      & $\checkmark$ & $\checkmark$ & $\checkmark$ & 95.23 \\
\hline
\end{tabular}
\label{tab:ablation}
\end{table}
In this section, we evaluate four encoding strategies on both image classification and object detection tasks. For classification, we use the high performance SNN architecture QKFormer-3-256 \cite{23}. For object detection, we develop a spiking version of YOLOv5n \cite{25}, using SEW-ResNet blocks for residual connections to ensure fair comparison across encoding schemes \cite{12}. All experiments are conducted under an extremely low time-step regime (\( T = 4 \)) to evaluate efficiency in ultra short temporal windows. Additional model configurations are provided in the~\ref{app3}.

\subsection{Ablation Study}
\label{subsec1}

Using TTFS as a representative case, we observe that most of its performance gain arises simply from replacing raw image encoding with convolutional features. This improvement is largely attributable to the substantial channel expansion in the first convolutional layer, which compensates for the shallow temporal depth (T=4). 

In contrast, surrogate gradients are ineffective when applied directly to raw images, as no learnable parameters precede spike generation. When combined with a learnable convolution, surrogate gradients offer only a modest additional gain (\(\approx\)1 percentage point), as shown in Table~\ref{tab:ablation}. 

For reference, rate and phase encoding achieve 86.67\% and 87.77\% top-1 accuracy under the same four step setting, further confirming that TTFS’s performance gain mainly comes from convolutional feature extraction.

\subsection{Validation Experiments} 
\label{subsec1} 

The results show that all four encoding methods achieve comparable performance on image classification tasks, where temporal modeling plays a limited role under static inputs.In contrast, the performance differences become more evident on the detection task.
Compared with image classification, detection poses a more complex prediction setting, which makes the model more sensitive to the availability of temporal variation. As a result, the proposed phase-based encoding shows a clear advantage, while direct and rate encoding remain limited due to the absence of meaningful temporal dynamics. The results across datasets are summarized in Table~\ref{tab:encoding_comparison}.

\begin{table}[b]
\centering
\caption{
Performance comparison of four encoding schemes on CIFAR-10, CIFAR-100, and VOC.
For fairness, all methods share the same convolutional front-end, consisting of the learnability
of the convolution layer and surrogate gradient, as direct encoding inherently relies on convolutional learnability; learnable thresholds are also included 
for consistency. 
}
\setlength{\tabcolsep}{4pt}
\small
\renewcommand{\arraystretch}{1.1}
\begin{tabular}{lccc}
\toprule
\textbf{Encoding} & \textbf{CIFAR-10 Acc@1} & \textbf{CIFAR-100 Acc@1} & \textbf{VOC mAP@0.5} \\
\midrule
Direct           & 95.60 & 77.90 & 0.452 \\
Rate+Conv        & 95.62 & 78.08 & 0.444 \\
Phase+Conv       & \textbf{95.64} & \textbf{78.56} & \textbf{0.467} \\
TTFS+Conv        & 95.26 & 77.93 & 0.460 \\
\bottomrule

\end{tabular}
\label{tab:encoding_comparison}
\end{table}

\subsection{Discussion}


To illustrate how channel redundancy affects the role of surrogate gradients, we consider a setting where the first convolutional layer is restricted to three output channels under a TTFS like firing scheme. In this case, performance increases from 83.22\% to 86.43\% when a learnable convolution is used, but increases to 93.71\% when surrogate gradients are used in conjunction with the learnable convolution.

This example demonstrates that the benefit of learnable convolution relies on sufficient channel expansion in the first layer. When the channel capacity is limited, the influence of surrogate gradients becomes significantly more pronounced, revealing their essential contribution beyond the encoding scheme.

Moreover, different encoding schemes exhibit distinct behaviors across classification and detection tasks. We argue that this discrepancy likely does not stem from the encoding schemes themselves, but rather from whether the network actually requires additional temporal depth for effective representation. Using direct encoding as an example, reducing the time steps in QKFormer from four to two yields only a marginal drop on CIFAR-10 (95.6\%→94.89\%), yet causes a noticeably larger decline on VOC detection (mAP 0.452→0.365).

This suggests that modern SNN backbones can solve coarse grained classification with minimal reliance on temporal structure, whereas fine-grained localization tasks remain more sensitive to reductions in temporal integration.


\section{Conclusion and Future Work}



This work re-examines direct encoding in static-image SNNs, demonstrating its functional equivalence to rate coding and showing that the reported performance gaps mainly arise from the learnability of convolutional layers and the use of surrogate gradients. To address the lack of temporal dynamics, we introduce a minimal learnable temporal encoding that adds adaptive phase shifts to generate meaningful temporal variation.

Experiments verify that this temporal encoding restores useful time structure and improves performance on tasks that rely on temporal processing. While this study provides a unified perspective on three mainstream encoding paradigms, their practical advantages may vary across scenarios for instance, TTFS based methods may be preferable for ultra low power applications, whereas rate based approaches may offer greater robustness in noisy environments.

Overall, our findings highlight the necessity of explicit temporal structure when static inputs contain no inherent time variation. Future work will explore broader integration of temporal design principles with other components of SNNs to enhance learning under static-input conditions.


\appendix
\section{Experimental Results with LIF Neurons}
\label{app1}
\setcounter{table}{0}
We compare standard LIF neurons with learnable LIF neurons to investigate the impact of neuron model variations on encoding efficacy. The baseline ANN (YOLOv5n) achieves an mAP@.5 of 0.668 on VOC. As shown in Table \ref{tab:lif}, standard LIF neurons show minimal performance differences across encoding strategies.

\begin{table}[h]
\centering
\caption{Experimental results with LIF neurons.}
\label{tab:lif}
\small
\begin{tabular}{lccc}
\toprule
\textbf{Encoding} & \textbf{CIFAR-10 Acc@1} & \textbf{CIFAR-100 Acc@1} & \textbf{VOC mAP@.5} \\
\midrule
Direct Encoding   & 95.37                   & 78.35                    & 0.451               \\
Rate Encoding     & 95.46                   & 78.08                    & 0.442               \\
Phase Encoding    & 95.56                   & 77.71                    & 0.451               \\
TTFS              & 95.23                   & 77.82                    & 0.439               \\
\bottomrule
\end{tabular}
\end{table}

\section{Concrete Example of Direct Encoding to Rate Coding Equivalence}
\label{app2}
To intuitively illustrate the rate-coding equivalence of direct encoding under static inputs, we consider the LIF neuron model with $T = 4$ time steps, membrane decay constant $\tau = 0.5$, and soft reset (membrane potential subtracts 1 upon firing). When input current $X$ is constant across time steps, it uniquely maps to a firing frequency, forming rate coding. Only seven distinct firing patterns are possible, as listed in Table B.1.

Furthermore, we conduct a \textbf{temporal shuffling experiment} to assess reliance on temporal ordering by randomly permuting spike trains across time steps while preserving the total number of spikes per neuron. Minimal performance degradation suggests that the network relies on firing rates rather than precise temporal patterns.


On CIFAR-10, accuracy drops slightly from 95.37\% to 95.01\% after shuffling, confirming that information is encoded via spike rates. 

\setcounter{table}{0}
\begin{table}[h]
\centering
\caption{Firing patterns for constant input current \( X \) over four time steps.}
\label{tab:rate-patterns}
\begin{tabular}{ccc}
\toprule
\textbf{Firing Pattern} & \textbf{Boundary Range} \\
\midrule
0000 & \( X < 1.066 \) \\
0001 & \( 1.066 \leq X < 1.142 \) \\
0010 & \( 1.142 \leq X < 1.333 \) \\
0101 & \( 1.333 \leq X < 1.714 \) \\
0110 & \( 1.714 \leq X < 1.866 \) \\
0111 & \( 1.866 \leq X < 2 \) \\
1111 & \( 2 \leq X \) \\
\bottomrule
\end{tabular}
\end{table}


\section{Experimental Settings}
\label{app3}
To ensure reproducibility, we adopt the following unified experimental settings: time steps $T = 4$; LIF neuron membrane decay constant $\tau = 0.5$; surrogate gradient: Sigmoid; pretrained weights: none; initial threshold values: 1. For CIFAR tasks, we use QKFormer-3-256 with parameters consistent with the original implementation. For object detection, we employ Spiking-YOLOv5n with 300 training epochs and batch size 24. Random seeds are set to 35, 1000, and 0 (three runs). To balance computational cost and repeatability, detection experiments use VOC images resized to $320 \times 320$. This setting does not alter the relative performance trends across encoding schemes, affecting only absolute accuracy levels. All other hyperparameters follow the public source code. Experiments are conducted using PyTorch on a single-GPU environment.






\bibliographystyle{elsarticle-num}  
\bibliography{refer}  
\end{document}